\documentclass[11pt]{article}
\usepackage[left=2cm, right=2cm, top=1in, bottom=1in]{geometry}
\usepackage[hyphens]{url}
\usepackage[utf8]{inputenc}
\usepackage[T1]{fontenc} 
\usepackage{url}
\usepackage{mathtools}
\usepackage{amsmath}
\usepackage{amsfonts}
\usepackage{algorithm}
\usepackage{algorithmicx}
\usepackage{setspace}
\usepackage{xcolor}

\DeclareUnicodeCharacter{0301}{\'{e}}

\def\R{\mathbb R}

\DeclarePairedDelimiter{\norm}{\lVert}{\rVert}

\DeclareMathOperator*{\argmax}{Argmax}

\algrenewcommand\alglinenumber[1]{\footnotesize #1}

\usepackage{nicematrix}
\usepackage{booktabs} % for professional tables

\usepackage{biblatex}
\title{Laplacian-based Semi-Supervised Learning in\\ Multilayer Hypergraphs by Coordinate Descent}
\author{Sara Venturini\thanks{Department of Mathematics “Tullio Levi-Civita”, University of Padova, 35121 Padova, Italy (sara.venturini@math.unipd.it)} , 
Andrea Cristofari\thanks{Department of Civil Engineering and Computer Science Engineering, University of Rome ``Tor Vergata'', 00133 Rome, Italy (andrea.cristofari@uniroma2.it)},
Francesco Rinaldi\thanks{Department of Mathematics “Tullio Levi-Civita”, University of Padova, 35121 Padova, Italy (rinaldi@math.unipd.it)},
Francesco Tudisco\thanks{School of Mathematics, Gran Sasso Science Institute, Italy, (francesco.tudisco@gssi.it)}}
\date{\vspace{-5ex}}

\begin{document}

\maketitle
%\blfootnote{A. Cristofari, F. Rinaldi and F. Tudisco contributed equally to this work.}
\textbf{Abstract.}
Graph Semi-Supervised learning is an important data analysis tool, where given a graph and a set of labeled nodes, the aim is to infer the labels to the remaining unlabeled nodes.
In this paper, we start by considering an optimization-based formulation of the problem for an undirected graph, and then we extend this formulation to  multilayer hypergraphs. 
We solve the problem using different coordinate descent approaches and compare the results with the ones obtained by the classic gradient descent method.
Experiments on synthetic and real-world datasets show the potential of using coordinate descent methods with suitable selection rules.\\
\textbf{Key words.}
semi-supervised learning, coordinate methods, multilayer hypergraphs.\\

\section{Introduction}
Consider a finite, weighted and undirected graph $G = (V,E,w)$, with node set $V$, edge set $E \subseteq V\times V$ and edge-weight function $w$ such that $w(e)=w(uv)>0$ if $e = (u,v)\in E$ and $0$ otherwise. 
%\new{We assume that is given a symmetric weight matrix $W \in \mathbb{R^{|V|\times |V|}}$ on the edges.}
Suppose each node $u\in V$ can be assigned to one of $m$ classes, or labels, $C_1,\dots,C_m$. 
In graph-based Semi-Supervised Learning (SSL), given a graph  $G$ and an observation set of labeled nodes $O\subset V$ whose vertices $u\in O$ are pre-assigned to some label $y_u\in \{C_1,\dots,C_m\}$, the aim is to infer the labels of the remaining unlabeled nodes in $V\setminus O$, using the information encoded by the graph \cite{chapelle2010semi,song2022graph,subramanya2014graph}.
 % is chosen so that the weight of an edge  between two nodes measures the similarity between them.
% Each vertex $u$ in the observation set $O \subset V$ is assigned a label $y_u$. %, where $y : O \rightarrow \mathbb{R}^h$.

Extending labels is a-priori an ill-posed problem since there are infinitely many solutions. Therefore, a common approach is to proceed by  making the so-called semi-supervised smoothness assumption. This assumption requires that good labeling functions $z_j: V \rightarrow \mathbb{R_+}$ for the $j$-th class, whose entries $z_{u,j}$ quantify the likelihood that $u\in V\setminus O$ belongs to $C_j$,  % should  which requires  that the learned labeling function $z : V \rightarrow \mathbb{R}$
should be smooth in densely connected regions of the graph.
Assuming that the edges of the graph represent some form of similarity between pairs of nodes, this smoothness assumption corresponds to assuming  that similar nodes are likely to have similar labels.

Consider the following  $\ell_{2}$-based Laplacian regularizer \cite{zhu2003semi}  % In order to force similar nodes in dense regions of the graph to have similar labels, it is successful to use the $\ell_{2}$-based Laplacian regularizer \cite{zhu2003semi}
\begin{equation}
r_2(z) = \frac{1}{2} \sum_{(u,v) \in E}  w(uv) ( z_u - z_v )^2\, .
\end{equation}
Minimizing $r_2(z)$ subject to either hard label
constraints, $z_u = y_u$ for $u \in O$, or a soft penalty constraint like the mean squared error $\sum_u ( y_u - z_u )^2$, with respect to the known labels $y$, is a successful way to enforce smoothness with respect to the edges. 
In both cases, the resulting objective function is strictly convex and hence the corresponding minimization problem has a unique optimal solution.

Even if the $\ell_{2}$-based Laplacian regularizer is very popular and effective in many situations, it has been proved that it can yield degenerate solutions in the presence of very few input labels in $O$, because the learned function $z$ becomes nearly constant on the whole graph, with sharp spikes near the labeled data $O$ %, in order to respect the measurement constraints 
\cite{el2016asymptotic,nadler2009semi}.
Therefore, several alternative formulations have been proposed \cite{kyng2015algorithms,zhou2011semi}, including approaches based on total variation \cite{cristofari2020total, hein2013total} and the class of $p$-Laplacian based regularizers \cite{el2016asymptotic}, more in general, defined as  
\begin{equation}
r_p(z) = \frac{1}{p} \sum_{(u,v) \in E} w(uv) | z_u - z_v |^p\, .
\end{equation}
Note that this objective function is still strictly convex. Moreover, $r_p$ discourages the solution from developing sharp spikes for $p > 2$, giving a heavier penalty on large gradients $| z_u - z_v |$. 
Choosing instead $1 \leq p < 2$ encourages the gradient to be sparse. Furthermore, when $p\to 1$, the resulting objective function is  directly connected with graph cuts and modular  clusters \cite{buhler2009spectral,tudisco2018community,tudisco2022duality}.
Many works studied the behaviour of $r_p$ as $p$ varies, mainly for graphs generated by the geometric random graph model \cite{calder2018game,el2016asymptotic, flores2022analysis, slepcev2019analysis,tudisco2016nodal}. 

In this paper, we want to investigate the effectiveness of these types of Laplacian regularizers for the task of  graph semi-supervised learning, but taking into consideration also higher-order interactions. 
A variety of complex systems has been successfully described as networks whose interacting pairs of nodes are connected by links. However, in real-world applications, we need to describe interactions in more detailed and varied ways \cite{battiston2020networks,boccaletti2014structure}. 
On the one hand, we have simplicial complexes or hypergraphs, which are the natural candidates to describe collective actions of groups of nodes \cite{chitra2019random,ibrahim2020local,veldt2020minimizing,yin2017local,zhang2017re}. On the other hand, we have  multilayer networks, i.e., networks that are coupled to each other through different layers, all of them representing different type of relationships between the nodes \cite{argyriou2005combining,eswaran2017zoobp,gujral2018smacd,karasuyama2013multiple,mercado2019generalized,nie2016parameter,tsuda2005fast,zhou2007spectral}. Evidence shows that each of those tools can improve modeling capacities with respect to standard graphs.
Multilayer hypergraphs arise naturally in diverse applications such as  science of science (e.g., nodes represent authors and in one layer a group of authors is an hyperedge if they wrote a paper together, while, in another layer, pairs of nodes are connected if they cite each other),  protein networks (e.g., nodes are proteins and they can be connected in pairs or in groups using multiple complementary genomic data which are the different layers),  social networks (e.g., nodes are users and they can interact in groups using different platforms).
Here, we focus on multiplex hypergraphs, modeled by a sequence of hypergraphs (the layers) with a common set of nodes and no hyperedges between nodes of different layers. 
Moreover, with the terminology introduced in~\cite{magnani2021community} in the context of multilayer networks, 
our aim is to find a set of communities that is \textit{total} (i.e., every node belongs to at least one community),
\textit{node-disjoint} (i.e., no node belongs to more than one cluster on a single layer), and \textit{pillar} (i.e., each node belongs to the same community across the layers).

However, relatively few studies have  considered both multilayer and higher-order structures in complex networks so far \cite{whang2020mega}.
This is mainly due to the fact that getting a good solution for models with such a complex structure comes at a much higher computational cost. In this work, we hence take a first step in the study of semi-supervised learning over multilayer hypergraphs, trying to deal with this additional complexity.

We solve the problem using different coordinate descent approaches and compare the results with the ones obtained by classic first-order approaches, like, e.g., gradient descent/label spreading.
Even though coordinate descent approaches were used in the literature to deal with other semi-supervised learning problems \cite{demiriz2001optimization,dong2018learning}, the analysis reported in this paper represents, to the best of our knowledge, the first attempt to give a thorough analysis of those methods for semi-supervised learning in multilayer hypergraphs. 

The rest of the paper is organized as follows.
In Section~\ref{Problem statement}, we introduce the graph semi-supervised problem and the formulation for multilayer hypergraphs. 
In Section~\ref{sec:bcd}, we briefly review
the block coordinate descent approaches.
In Section~\ref{Numerical experiments}, we report the results of experiments on synthetic and real-world datasets. In 
Section~\ref{Conclusions}, we draw some conclusions.
Finally, in~\ref{appendix:A}, we report some computations needed to apply the methods to our specific problem, pointing out the differences in the special case $p=2$. 

%\section{The optimization problem}
\section{Problem statement}
\label{Problem statement}
In this section, we formalize the notation and formulate the problem under analysis.
Consider first an undirected and weighted graph $G = (V,E,w)$ with node set $V$ and edge set $E$. Let  $A = (A_{uv})_{u,v \in V}$ be the adjacency matrix of $G$, with weights $A_{uv}=w(e)>0$ for $e=(u,v)\in E$, measuring the strength of the tie between nodes $u$ and $v$, and $A_{uv}=0$ if $(u,v)\notin E$.
We assume that $V$ can be partitioned into $m$ classes $C_1, \ldots, C_m$ and that, only for a few nodes in $O\subset V$, it is known the class $C_j$ to which they belong. The problem consists in assigning the remaining nodes to a class.

%In the literature, different approaches have been proposed to solve this problem. [...]

Here, we review  the approach based on the $p$-Laplacian regularization and the corresponding  optimization problem. Define the $(|V| \times m)$-dimensional matrix of the input labels $Y$, such that
\[
Y_{u,j} =
\begin{cases}
\dfrac1{|C_j\cap O|} & \quad \text{if node } u\in O \text{ belongs to the class } C_j, \\[2ex]
0 & \quad \text{otherwise},
\end{cases}
\]
where $|C_j\cap O|$ is the cardinality of the known class $C_j$, i.e., the number of nodes that are initially known to belong to $C_j$.
Now, let $y^j$ be the $j$th column of $Y$ and, for all $u \in V$,
let $\delta_u$ be the weighted degree of $u$, that is,
$\delta_u = \sum_{v \in V} A_{uv}$. 
%
% Then, for all class $j \in \{1,\ldots,k\}$, we want to solve the following problem:
% \begin{equation}\label{prob_graph}
% \min_{z \in \R^{|V|}} \norm{z-y^j}^2 + \lambda \sum_{u,v \in V} A_{uv} \biggl(\frac{z_u}{\sqrt{\delta_u}}-\frac{z_v}{\sqrt{\delta_v}}\biggr)^2,
% \end{equation}
% with a given regularization parameter $\lambda \ge 0$.
% After computing the optimal solutions $(z^*)^j$ of the above problems for all $j \in \{1,\ldots,k\}$, we can interpret the $i$th entry of $(z^*)^j$ as the probability that the node $i \in V$ belongs to the class $C_j$. So, it is natural to assign each node $i \in V$ to the class $j \in \argmax_{r=1,\ldots,k} (z^*)^r_i$.
%
The Laplacian regularized SSL problem boils down to the following minimization problem for all classes $j \in \{1,\ldots,m\}$:
%to minimize the following function over $z \in \R^{|V|}$:
\begin{equation}\label{prob_graph}
%f_{j}(z) = 
\min_{z \in \R^{|V|}} \norm{z-y^j}^2 + \lambda \sum_{u,v \in V} A_{uv} \biggl|\frac{z_u}{\sqrt{\delta_u}}-\frac{z_v}{\sqrt{\delta_v}}\biggr|^p,
\end{equation}
with given $p\geq 1$ and regularization parameter $\lambda \ge 0$. 
%The minimization of the functions $f_{1},\ldots,f_{k}$, is equivalent to minimize the following function over $Z \in \R^{|V| \times k}$:
Equivalently, as the minimization problems above are independent for $j \in \{1,\ldots,m\}$, we can simultaneously optimize their sum, which can be written in compact matrix notation as %obtaining the matrix , The minimization of the above problems for $j \in \{1,\ldots,m\}$, is equivalent to the following one:
\begin{equation} \label{prob_graph_matrix}
%F(Z) =  
\min_{Z \in \R^{|V| \times m}} \lVert  Z -Y \rVert_{(2)}^2 +  \lambda \lVert W^{1/p} B D^{-1/2}Z \rVert_{(p)}^p,
\end{equation}
where $\|M\|_{(p)}$ denotes the entry-wise $\ell^p$ norm of the matrix $M$, $D$ is the $|V|\times |V|$ diagonal matrix of the graph degrees 
\begin{equation*}
D = 
\begin{bmatrix}
\delta_{1} & 0 & \cdots & 0 \\
0 & \delta_{2} & \cdots & 0 \\
\vdots  & \vdots  & \ddots & \vdots  \\
0 & 0 & \cdots & \delta_{|V|} 
\end{bmatrix}\, ,
\end{equation*} 
%$\lVert \cdot \rVert_{2}^2$ is the Frobenius norm,
%$\lVert M \rVert_{cl(p)}^{p} = \sum_{h \text{ column of} M} \lVert h\rVert_{p}^p$, 
 $B$ is the $|E|\times |V|$ (signed) incidence matrix of the graph, which for any chosen orientation of the edges is entrywise defined as 
\[
B_{e,u} =
\begin{cases}
1 & \quad \text{if node } u \text{ is the source of edge } e, \\
-1 & \quad \text{if node } u \text{ is the tip of edge } e, \\
0 & \quad \text{otherwise},
\end{cases}
\]
 and $W$ is the diagonal $|E|\times |E|$ matrix of the edge weights $W_{e,e}=w(e)$. Note that, even though we are dealing with undirected graphs, $B$ requires fixing an orientation for the edges of $G$. However, all the arguments presented here are independent of the chosen orientation.
 % ....CENTRATA... $B_{u,e} = 1$ if $u$ is the source of $e$, $B_{u,e}=-1$ if $u$ is the tip of $e$, and $B_{u,e}=0$ otherwise.... 
 % \textcolor{red}{Chiarire che questa e'quella segnata}. %, and
% \begin{equation*}
% D^{-\frac{1}{2}} = 
% \begin{bmatrix}
% \frac{1}{\sqrt{\delta_{1}}} & 0 & \cdots & 0 \\
% 0 & \frac{1}{\sqrt{\delta_{2}}} & \cdots & 0 \\
% \vdots  & \vdots  & \ddots & \vdots  \\
% 0 & 0 & \cdots & \frac{1}{\sqrt{\delta_{|V|}}} 
% \end{bmatrix}
% \end{equation*}
For $p=2$, a direct computation shows that the 
the optimal solution $Z^*$ of the above problem is entrywise nonnegative. The same property carries over to any $p\geq 1$,  as one can interpret the minimizer of \eqref{prob_graph_matrix} as the smallest solution of a $p$-Laplacian eigenvalue equation on $G$ with boundary conditions, see e.g. \cite{deidda2023nodal}. Thus, we can interpret the entry $Z^*_{u,j}\geq 0$ as a score that quantifies how likely it is for the node $u \in V$ to belong to the class $C_j$ and we then assign each node $u \in V$ to the class $j \in \argmax_{r=1,\ldots,m} Z^*_{u,r}$.

% Now, we want to extend formulation~\eqref{prob_graph} to the case where $G$ is a multilayer hypergraph.
% Specifically, assume that we have $L$ layers $G^1, \ldots, G^L$, where $G^{\ell}(V,E^{\ell})$ is the graph forming the $\ell$th layer and $E_{\ell}$ is a hyperedge set,
% that is, $E^{\ell}$ contains interactions of order greater than 2.
% First, we represent each $G^{\ell}(V,E^{\ell})$ via a clique-expanded graph, given by the adjacency matrix
% \[
% A^{\ell} = I^{\ell}I^{\ell T} - D^{\ell},
% \]
% with $I^{\ell} $ being the incident matrix of the hypergraph and $D^{\ell}$ being the degree diagonal matrix. Accordingly, $\delta^{\ell}_u$ is the weighted degree of a node $u \in V$, that is, $\delta^{\ell}_u = \sum_{v \in V} A^{\ell}_{uv}$.

Now, we want to extend the formulation~\eqref{prob_graph_matrix} to the case where rather than a graph $G$, we have a multilayer hypergraph $H$.
Specifically, assume that we have $L$ layers $H_1, \ldots, H_L$, where $H_{\ell}(V,E_{\ell})$ is the hypergraph forming the $\ell$th layer and $E_{\ell}$ is a hyperedge set,
that is, $E_{\ell}$ contains interactions of order greater than 2. In other words, each $e\in E_\ell$ is a set of  arbitrary many nodes, weighted by $w_\ell(e)>0$. 
The topological information of a hypergraph $H_{\ell}$ can be all included in the (signless) incidence matrix $K_{\ell} \in \R^{|E_{\ell}|\times |V|}$, defined as $(K_\ell)_{e,u}=1$ if $u\in e$, and $(K_\ell)_{e,u}=0$ if $u\not\in e$, for all $u\in V$ and $e\in E_\ell$, see e.g. \cite{battiston2020networks,prokopchik2022nonlinear,zhou2006learning}. %. where the nonzero entries represent the number of times vertex $u$ is present in the relative hyperedge 
Using $K_\ell$, we can represent each $H_{\ell}(V,E_{\ell})$ via a clique-expanded graph $G(H_\ell)$, which corresponds to the adjacency matrix
\[
A_{\ell} = K_{\ell}^T W_\ell K_{\ell} - D_{\ell},
\]
with $W_\ell$ being the $|E_\ell|\times |E_\ell|$ diagonal matrix of the relative hyperedge weights, defined as 
\[
(W_\ell)_{e,e}=\frac {w_\ell(e)}{|e|}>0,
\] 
and $D_{\ell}$ being the diagonal matrix of the node degrees of the hypergraph $H_\ell$, defined as 
\[(D_\ell)_{u,u}=(\delta_\ell)_u = \sum_{e\in E}w_\ell(e)|e|^{-1}(K_\ell)_{e,u} = (K_{\ell}^T W_\ell K_{\ell})_{u,u}\, .
\] 
Note that the edge $(u,v)$ is in  the resulting clique-expanded graph $G(H_\ell)$ if and only if $u\neq v$ and there exists at least one hyperedge in $E_\ell$ such that both $u\in E_\ell$ and $v\in E_\ell$. In that case, the weight of the edge $(u,v)$ in $G(H_\ell)$ is 
\[
(A_\ell)_{u,v}=\sum_{e: u,v\in e}\frac{w_\ell(e)}{|e|}
\]
and we have $(\delta_\ell)_u=\sum_{v\in V}(A_\ell)_{u,v}$. Therefore, proceeding as before,  we can define $B_\ell$ as the signed incidence matrix of $G(H_\ell)$ and we can sum the corresponding regularization terms across all the layers, obtaining the following formulation:
\begin{align}
\label{prob_hypermultilayergraph}
    \min_{Z \in \R^{|V| \times m}} &\vartheta (Z) := f(Z)+r_p(Z)\\
    %& f(Z) = F(Z)+G(Z) \\
    \notag\text{where } & f(Z) = \|Z-Y\|_{(2)}^2, \quad r_p(Z) = \sum_{\ell=1}^L \lambda_{\ell} \lVert W_\ell^{1/2} B_{\ell} D_{\ell}^{-1/2}Z \rVert_{(p)}^p,
\end{align}
% \begin{equation}\label{prob_hypermultilayergraph}
% \min_{Z \in \R^{|V| \times k}} \lVert  Z -Y \rVert_{(2)}^2 +  \sum_{\ell=1}^L \lambda^{\ell} \lVert B^{\ell} D^{^\ell-\frac{1}{2}}Z \rVert_{(p)}^p 
% \end{equation}
where $\lambda_1, \ldots, \lambda_L \ge 0$ are regularization parameters. Note that, if $H$ is a standard graph, i.e., if $|e|=2$ for all edges and $L=1$, then \eqref{prob_hypermultilayergraph} boils down to \eqref{prob_graph_matrix}, up to the constant term $1/|e|=1/2$. Note moreover that, as in the graph case, we can equivalently write the objective function $\vartheta(Z)$ as $\sum_j \vartheta_j(z^j)$, where $z^j$ is the $j$-th column of $Z$, and 
\[
\vartheta_j(z) = \|z-y^j\|_2^2 + \sum_{\ell=1}^L \lambda_\ell \sum_{e\in E_\ell} \frac{w_\ell(e)}{|e|} \sum_{u,v\in e}\left| \frac{z_u}{\sqrt{(\delta_\ell)_u}} -\frac{z_v}{\sqrt{(\delta_\ell)_v}}  \right|^p .
\]
%We will refer to the function that we want to minimize in~\eqref{prob_hypermultilayergraph} as $F_p(Z)$.
The above expression shows that the regularizers $\vartheta_j$ enforce a form of higher-order smoothness assumption in the solution across all the nodes of each layer's hyperedge by imposing the minimizer $Z^*$ to have similar values on pairs of nodes in the same hyperedge.  This immediately justifies the choice of the objective function \eqref{prob_hypermultilayergraph} for SSL on multilayer hypergraphs.
Also note that, as in the graph setting, the  optimal solution $Z^*$ to \eqref{prob_hypermultilayergraph} has to be entrywise nonnegative and thus, once $Z^*$ is computed, we can assign each node $u \in V$ to the class $j \in \argmax_{r=1,\ldots,m} Z^*_{u,r}$.

% Then, we can sum the regularization terms across the layers obtaining the following formulation for $j \in \{1,\ldots,k\}$:
% \begin{equation}\label{prob_hypermultilayergraph}
% \min_{z \in \R^{|V|}} \norm{z-y^j}^2 + \sum_{\ell=1}^L \lambda_{\ell} \sum_{u,v \in V} (A^{\ell})_{uv} \biggl(\frac{z_u}{\sqrt{\delta^{\ell}_u}}-\frac{z_v}{\sqrt{\delta^{\ell}_v}}\biggr)^2,
% \end{equation}
% with given regularization parameters $\lambda_1, \ldots, \lambda_L \ge 0$. As above, once an optimal solution $(z^*)^j$ of problem~\eqref{prob_hypermultilayergraph} is computed for all $j \in \{1,\ldots,k\}$, we can assign each node $i \in V$ to the class $j \in \argmax_{r=1,\ldots,k} (z^*)^r_i$.

\section{Block coordinate descent approaches}\label{sec:bcd}
When dealing with large-scale optimization problems, such as those arising in  semi-supervised learning problems on real-world multilayer hypergraphs,  traditional methods %for the solution of \eqref{prob_hypermultilayergraph}
may be impractical and block coordinate descent methods represent a valid tool to achieve high efficiency.
At every iteration of a block coordinate descent method, a working set of a few variables is suitably selected and properly updated,  while keeping the remaining variables fixed.
The general scheme for a block coordinate descent method to minimize an objective function $f(z)$ is reported in Algorithm~\ref{alg:bcd}.

\begin{algorithm}[h!]
\caption{Generic block coordinate descent method}
\label{alg:bcd}
\begin{algorithmic}[1]
\setcounter{ALG@line}{-1}
\setstretch{1.1}
\State \textbf{Given} $z^0 \in \R^n$
\State \textbf{For} $k=0,1,\ldots$
\State \hspace*{0.5truecm}Choose a working set $W^k \subseteq \{1,\ldots,n\}$
\State \hspace*{0.5truecm}Compute $s^k \in \R^n$ such that $s^k_i = 0$ for all $i \notin W^k$
\State \hspace*{0.5truecm}Set $z^{k+1} = z^k + s^k$
\State \textbf{End for}
\end{algorithmic}
\end{algorithm}

In the literature, many block coordinate descent methods were proposed for both unconstrained and constrained problems, differing from each other in the computation of $W^k$ and $s^k$ (see, e.g.,~\cite{wright:2015} and the references therein).
As for the computation of the working set $W^k$, a possible choice is to use a \textit{cyclic rule}, also known as \textit{Gauss-Seidel rule}~\cite{bertsekas:1999}. It consists in partitioning the variables into a number of blocks and selecting each of them in a cyclic fashion. This approach can be generalized to the \textit{essentially cyclic rule} or \textit{almost cyclic rule}~\cite{luo:1992}, requiring that each block of variables must be selected at least once within a prefixed number of iterations.
In unconstrained optimization, blocks can even be made of just one variable and every update (i.e., the computation of $s^k$) can be carried out by an exact or an inexact minimization~\cite{bertsekas:2015,bertsekas:1999,grippo:1999,luo:1992,sargent:1973}. These methods have been also extended to constrained settings, possibly requiring blocks being made of more than one variable when the constraints are not separable~\cite{bertsekas:1999,birgin:2022,cassioli:2013,cristofari:2019,grippo:2000,lucidi:2007,razaviyayn:2013}. A remarkable feature of cyclic based rules is that, at each iteration, only a few components of $\nabla f$ must be calculated.
This can lead to high efficiency when computing one component of $\nabla f$ is much cheaper than computing the whole gradient vector.

A second possibility to choose the working set is to use a \textit{random rule}, that is, $W^k$ can be computed randomly from a given probability distribution. These algorithms, usually known as \textit{random coordinate descent methods}, show nice convergence properties in expectation for both unconstrained~\cite{nesterov:2012,richtarik:2014} and constrained problems~\cite{ghaffari2022convergence,necoara:2017,necoara:2014,patrascu:2015,reddi:2015}. Note that random rules, as well as cyclic rules, do not use first-order information to compute the working set, thus still leading to high efficiency when the computation of one component of $\nabla f$ is much cheaper than the computation of the whole gradient vector.

Another way to choose the working set $W^k$ is to use a \textit{greedy rule}, also known as \textit{Gauss-Southwell rule}. It consists in selecting, at each iteration, a block containing the variable(s) that most violate a given optimality condition.  In the unconstrained case we can choose as working set, for instance, the block corresponding to the largest component of the gradient in absolute value.
Also for this rule, exact or inexact minimizations can be carried out to update the variables~\cite{cristofari:2023,de2016fast,grippo:1999,luo:1992} and extensions to constrained settings were considered in the literature~\cite{beck:2014,lin:2001,nutini:2022,tseng:2009a,tseng:2009b}. Generally speaking, a greedy rule might make more progress in the objective function, since it uses first (or higher) order information to choose the working set, but might be, in principle, more expensive than  cyclic or random selection. However, several recent works show that certain problem structures allow for efficient calculation of this class of rules in practice (see, e.g.,~\cite{nutini:2022} and references therein for further details).

\begin{algorithm}[h!]
\caption{Block coordinate descent method for problem~\eqref{prob_hypermultilayergraph} - matrix form}
\label{alg:bcd_matrix}
\begin{algorithmic}[1]
\setcounter{ALG@line}{-1}
\setstretch{1.1}
\State \textbf{Given} $Z^0 \in \R^{n \times m}$
\State \textbf{For} $k=0,1,\ldots$
\State \hspace*{0.5truecm}Choose a working set $W^k = W_1^k \times \ldots \times W_m^k \subseteq \{1, \dots , n\}^m$
\State \hspace*{0.5truecm}Compute $S^k \in \R^{n \times m}$ such that $S^k_{ij} = 0$ for all $i \notin W_j^k$
\State \hspace*{0.5truecm}Set $Z^{k+1} = Z^k + S^k$
\State \textbf{End for}
\end{algorithmic}
\end{algorithm}

In this work, we adapt block coordinate descent methods to solve problem~\eqref{prob_hypermultilayergraph}, leading to the method reported in Algorithm~\ref{alg:bcd_matrix}. 
In particular, we start with a matrix $Z^0 \in \R^{n \times m}$ and, at each iteration $k$, we choose a working set $W_j^k$ for each class $j \in \{1,\ldots,m\}$.
We highlight that problem~\eqref{prob_hypermultilayergraph} solves the same problem for the different classes $C_j$ with $j = 1,\ldots,m$ in a matrix form, but each of them is independent and can  eventually be solved in parallel.
% In this work, we adapt block coordinate descent methods to solve problem~\eqref{prob_hypermultilayergraph} using variants of the three block selection strategies described above to compute the working set, i.e.,  cyclic rule,  greedy rule and  random rule.

\subsection{Coordinate descent approaches}
\label{Coordinate descent approaches}
In this paper, we focus on  block coordinate descent approaches that use blocks $W^k_j$ of dimension 1, i.e., $W^k_j=\{i_j^k\}$, with $i_j^k$ being a variable index for class $j$ at iteration $k$. 
Then,  $Z^{k+1}$ is obtained by moving the variables $Z^k_{i_j^k j}$ along $-\nabla_{i_j^kj} \vartheta(Z^k)$ with a proper stepsize $\alpha_j^k$.
Namely, for any class $j \in \{1,\ldots,m\}$,
\begin{equation}\label{var_upd}
Z^{k+1}_{hj} =
\begin{cases}
Z^{k}_{hj} - \alpha_j^k \, \nabla_{hj} \vartheta(Z^k)\quad & \text{if } h = i_j^k, \\
Z^{k}_{hj} & \text{otherwise.}
\end{cases}
\end{equation}

Taking into account the possible choices described in Section~\ref{sec:bcd},  we consider the following algorithms:
\begin{itemize}
\item \textbf{Cyclic Coordinate Descent (CCD).} At every iteration $k$, a variable index $i^k\in \{1,\ldots,n\}$ is chosen in a cyclic fashion (i.e., by a Gauss-Seidel rule), and then $Z^{k+1}$ is obtained as in~\eqref{var_upd} by setting $i_j^k = i^k$ for all $j \in \{1,\ldots,m\}$.
A random permutation of the variables every $n$ iterations is also used, since it is known that this might lead to better practical performances in several cases (see, e.g., \cite{gurbuzbalaban:2020,wright:2015}).
\item \textbf{Random Coordinate Descent (RCD).} At every iteration $k$, a variable index $i^k \in \{1,\ldots,n\}$ is  randomly chosen from a uniform distribution, and then $Z^{k+1}$ is obtained as in~\eqref{var_upd} by setting $i_j^k = i^k$ for all $j \in \{1,\ldots,m\}$.
\item \textbf{Greedy Coordinate Descent (GCD).} At every iteration $k$, a variable index $i_j^k \in \{1,\ldots,n\}$ is chosen for every class $j \in \{1,\ldots,m\}$ as
\[
i_j^k \in \argmax_{i=1,\ldots,n} |\nabla_{ij} \vartheta(Z^k)|
\]
(i.e., by a Gauss-Southwell rule), and then $Z^{k+1}$ is obtained  as in~\eqref{var_upd}.
\end{itemize}

The GCD method guarantees good rates when proper conditions are met~\cite{nutini:2022}. Anyway,
since at each iteration we need to evaluate the whole gradient and search for
the best index in order to choose the block to be used in the update, it might
become very expensive when we tackle large-scale 
semi-supervised learning problems. %problems coming from semi-supervised learning problems. 
%To practically implement those methods,
To practically implement those methods, specific strategies hence need to be implemented. It is important to highlight that, due to the sparsity present in the semi-supervised learning problems we consider, it is possible to
implement the basic GCD rule in an efficient way  (by, e.g., tracking the gradient
element in a max-heap structure, using caching strategies), see, e.g., \cite{nutini:2022}.

Since the practical efficiency of coordinate methods strongly depends on how the algorithm is implemented,  we report, in~\ref{appendix:A}, details on the calculations needed to update the gradient of the objective functions at a given iteration.

\section{Numerical experiments}
\label{Numerical experiments}

First-order methods like, e.g., gradient descent/label spreading are widely used in the context of semi-supervised learning \cite{subramanya2014graph,tudisco2020nonlinear,NIPS2003_87682805}. This is the reason why
we compare the coordinate approaches described in Subsection~\ref{Coordinate descent approaches}, i.e., the Cyclic Coordinate Descent method (CCD), Random Coordinate Descent method (RCD) and  Greedy Coordinate Descent method (GCD), with the Gradient Descent (GD) in our experiments. 
In the first setting, when $p=2$, the objective function is quadratic and we used a stepsize depending on the coordinatewise Lipschitz constants (see, e.g., \cite{nesterov2012efficiency}). We highlight that while calculating the coordinatewise Lipschitz constants or a good upper bound is pretty straightforward in the considered case for coordinate approaches, the calculation of the global Lipschitz constant might get expensive for GD (especially when dealing with large-scale instances).
For the $p\neq2$ setting, for simplicity we used a stepsize depending on an upper bound of the Lipschitz constants for all the methods (see, e.g., \cite{karimi2016linear,nutini:2022, nutini2015coordinate}). The performance of the coordinate descent algorithms might of course be further improved by choosing a more sophisticated coordinate dependent stepsize strategy \cite{qu2016coordinate,richtarik2016distributed,richtarik2016parallel,salzo2022parallel}.
%In all the algorithms we used  a Lipschitz constant dependent stepsize (see, e.g., \cite{karimi2016linear,nutini:2022, nutini2015coordinate}). We highlight that while calculating the Lipschitz constant or a good upper bound is pretty straightforward in the considered case for coordinate approaches, such a calculation might get impractical for GD (especially when dealing with large-scale instances).
In order to show the advantages of using coordinate methods with respect to gradient descent-like approaches, and how efficient those methods are  in practice, we performed extensive experiments both on synthetic and real-world datasets.

We report the efficiency plots of the objective function and accuracy of the final partition (evaluated on the subset of unlabeled nodes). In our experiments, we use the number of flops (i.e., one-dimensional moves) as our measure of performance.
 Therefore, for a graph with $N$ nodes, the GD uses $N$ flops at each iteration (i.e.,  it changes all the $N$ components of the iterate), while the coordinate methods use just one flop per iteration (i.e., they change  just one component at the time). As already highlighted in \cite{nutini:2022}, 
this measure is far from perfect, especially when considering greedy methods, since it
ignores the computational cost of each iteration. However, it
gives an implementation- and problem-independent measure. Furthermore, in our case, it is easy to estimate
the cost per iteration (which is small when the strategy is suitably implemented). Thus, we will see how   a faster-converging method like GCD leads to a substantial performance gain on the considered application.
%**How report the results (flops)**
%flops not time
%For each flop: value objective function, accuracy (just on subset unlabeled nodes)

We fixed the regularization parameters at $\lambda_{\ell} = 1$ for $\ell=1,\ldots,L$ and we initialized the methods with $Z^0 = 0$.
We implemented all the methods using Matlab. We emphasize that the choice of the parameters $\lambda_\ell=1$ does not affect the performance analysis we carry out in this work and has been made to ensure a fair balance among all layers. In practice, the choice of these parameters may require a non-trivial parameter tuning phase which is typically either model- or data-based, see e.g. \cite{nie2016parameter,tsuda2005fast,venturini2023learning}. We implemented all the methods using Matlab.\footnote{Code and data are available at the GitHub page: \url{https://github.com/saraventurini/Semi-Supervised-Learning-in-Multilayer-Hypergraphs-by-Coordinate-Descent}.}

\subsection{Synthetic datasets}
\label{Synthetic datasets}
We generated synthetic datasets by means of the Stochastic Block Model (SBM)~\cite{holland1983stochastic}, a generative model for graphs with planted communities depending on suitably chosen parameters $p_{in}$ and $p_{out}$. Those parameters represent the edge probabilities: given nodes $u$ and $v$, the probability of observing an edge between them is $p_{in}$ (resp. $p_{out}$) if $u$ and $v$ belong to the same (resp. different) cluster.

Notice that solving problem~\eqref{prob_hypermultilayergraph} on a multilayer hypergraph is equivalent to solving the same problem over a simple graph with an adjacency matrix made by the weighted sum of the adjacency matrices of the clique-expanded graphs of each layer. Therefore, we generated single layer datasets by fixing $p_{in}=0.2$ and varying the ratio $p_{in}/p_{out} \in \{3.5,3,2.5,2\}$.
More precisely, we created networks with 4 communities of 125 nodes each. 
We tested the methods also considering different percentages $perc$ of known labels per community. In particular, we consider $perc \in \{3\%, 6\%, 9\%, 12\%\}$, i.e., respectively $3, 7, 11, 15$ known nodes per community.

\begin{figure*}[ht]
\centering
{\includegraphics[scale=0.9,trim = 1.2cm 0.5cm 23cm 0.5cm, clip]{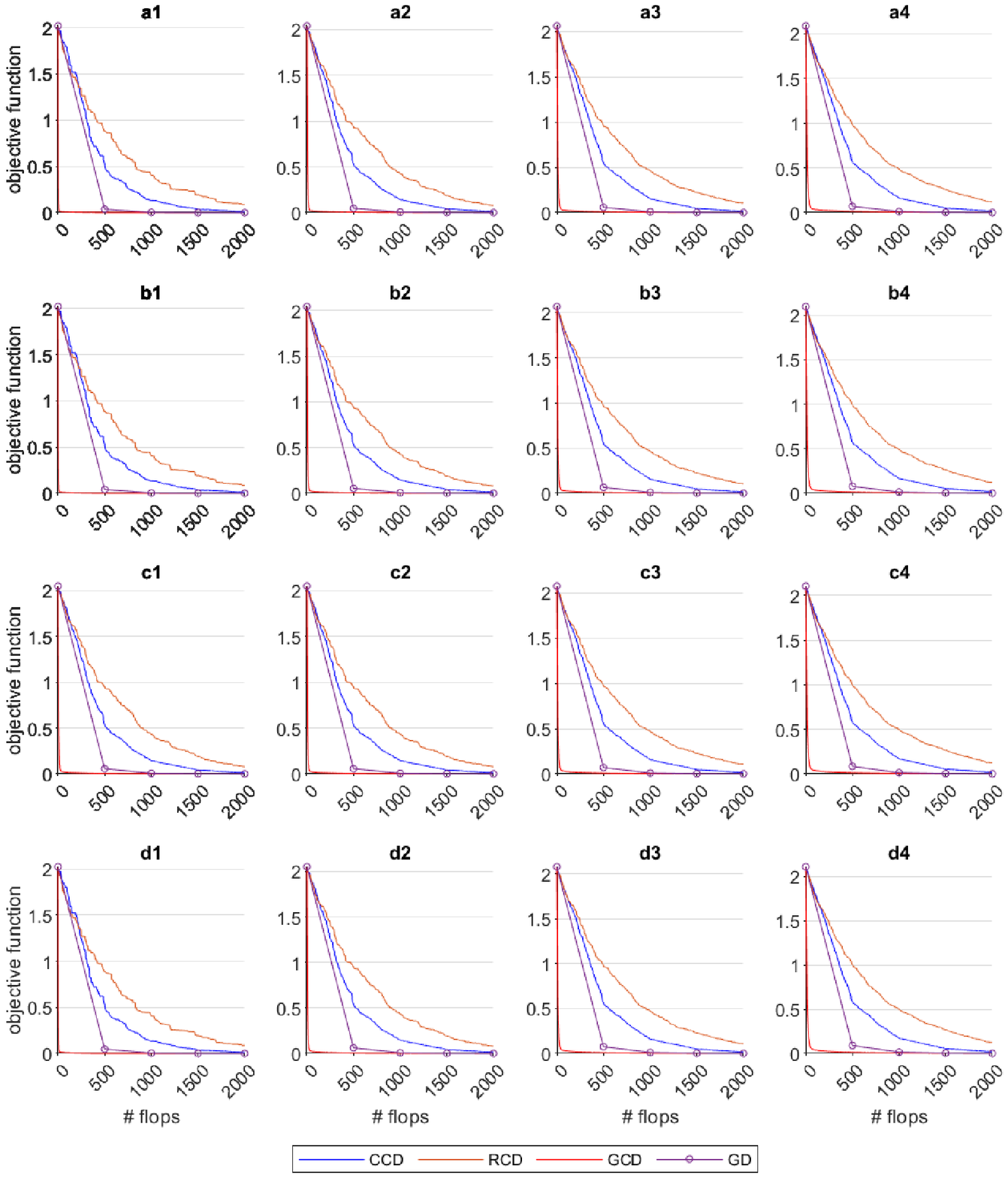}} 
\caption{Average values of the objective function over 5 random networks sampled from SBM for $p=2$, $p_{in}=0.2$, $\frac{p_{in}}{p_{out}} \in \{2,2.5,3,3.5\}$ varies in the rows and $perc \in [3\%, 6\%, 9\%, 12\%]$ varies in the columns.}
\label{fig:art_fun}
\end{figure*}

\begin{figure*}[ht]
\centering
{\includegraphics[scale=0.9,trim = 1.2cm 0.5cm 23cm 0.5cm, clip]{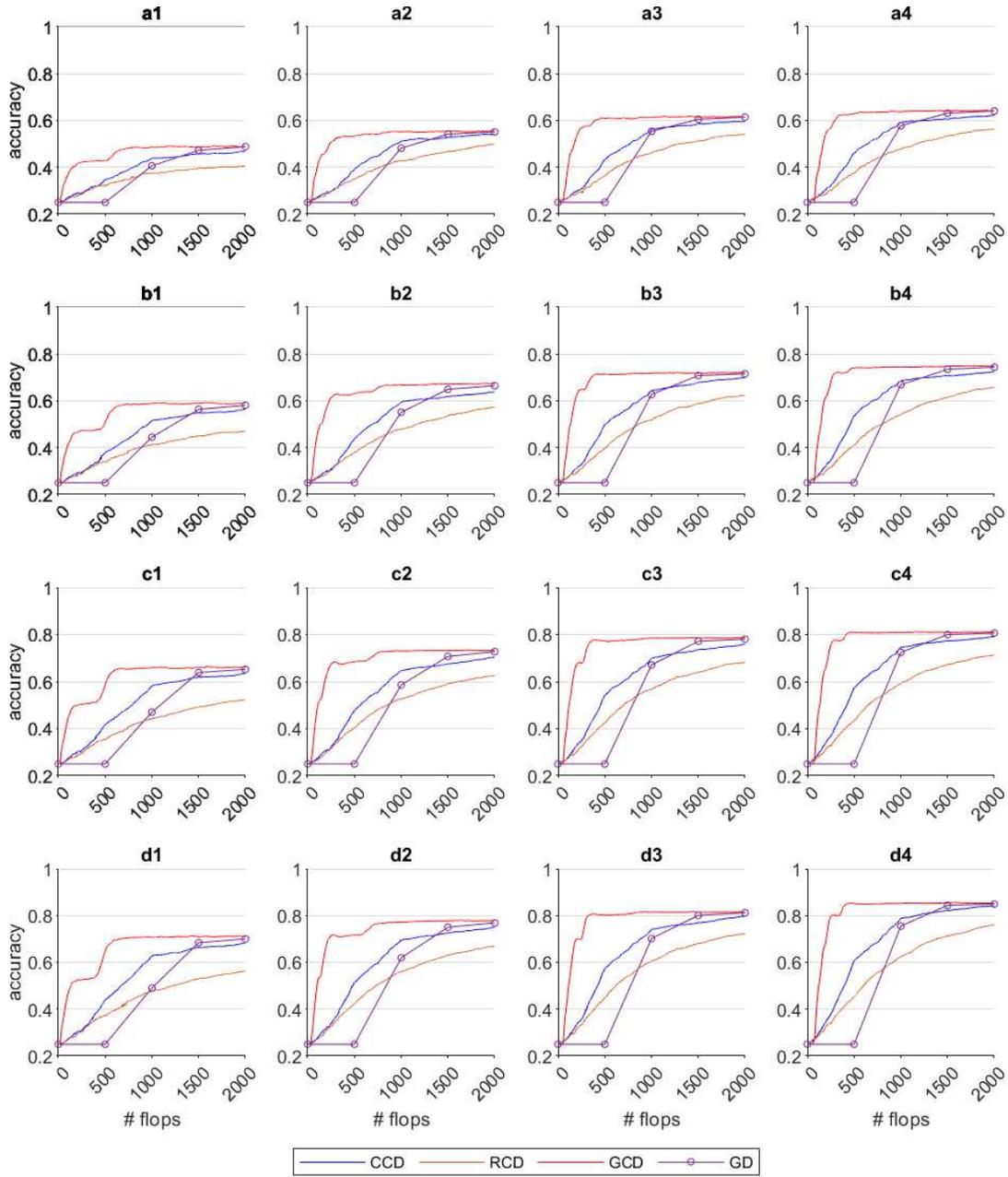}} 
\caption{Average values of the accuracy over 5 random networks sampled from SBM for $p=2$, $p_{in}=0.2$, $\frac{p_{in}}{p_{out}} \in \{2,2.5,3,3.5\}$ varies in the rows and $perc \in [3\%, 6\%, 9\%, 12\%]$ varies in the columns.}
\label{fig:art_acc}
\end{figure*}

We studied the optimization problem~\eqref{prob_hypermultilayergraph} fixing $p=2$. For each value of $(p_{out},perc)$ we sampled 5 random instances and considered average scores.
Results reported in Figures \ref{fig:art_fun} and \ref{fig:art_acc} respectively show the value of the objective function and the accuracy, in terms of number of flops. 
In each row, we report the results related to a fixed value of the ratio $p_{in}/p_{out}\in  \{2,2.5,3,3.5\}$ (ratio increasing top to bottom), varying the percentage of known labels $perc \in \{3\%, 6\%, 9\%, 12\%\}$ (percentage increasing left to right).
In Table~\ref{tab:art}, we present aggregated results of the objective function and the accuracy across the synthetic datasets (see Figures \ref{fig:art_fun} and \ref{fig:art_acc}).
For each method, it is shown average and standard deviation of the number of flops, normalized by the total number of nodes in the network, required to reach a certain level of objective/accuracy. This depends by a \textit{gate}, that is, a convergence tolerance, as in \cite{dolan2002benchmarking}.
It is also shown the fraction of failures, i.e., the fraction of problems where a method does not convergence within a number of iterations equal to 4 times the number of nodes. The averages are calculated without considering the failures and, in case of all failures, a hyphen is reported.
As we can easily see by taking a look at the plots and the tables, GCD always reaches a good solution in terms of both objective function value and accuracy, with a much lower number of flops than the other methods under consideration. 
Concerning the other coordinate methods under analysis, they seem to be slower than GD in getting a good  objective function, but faster in terms of accuracy. 
Therefore, if the coordinate selection is properly carried out, a coordinate method might outperform GD in practice. 

\setlength{\tabcolsep}{2.5pt}
\begin{table}[t]
\caption{
Aggregated results of the objective function (upper table) and the accuracy (lower table) across the synthetic datasets with $p=2$ (see  Figures \ref{fig:art_fun} and \ref{fig:art_acc}).
Using a tolerance \textit{gate}, for each algorithm \textit{flop} indicates the normalized number of flops (mean $\pm$ standard deviation) and \textit{fail} indicates the fraction of failures (i.e., stopping criterion not satisfied within the maximum number of iterations, set equal to 4 times the number of nodes). The averages are calculated without considering the failures and, in case of all failures, a hyphen is reported.}
\label{tab:art}
    \centering
    \begin{scriptsize}
    \vskip 0.15in
    \begin{NiceTabular}{l cc cc cc cc}
    \toprule
        & \multicolumn{2}{c}{CCD} & \multicolumn{2}{c}{RCD} & \multicolumn{2}{c}{GCD} & \multicolumn{2}{c}{GD}\\ 
        \cmidrule(lr){2-3} \cmidrule(lr){4-5} \cmidrule(lr){6-7} \cmidrule(lr){8-9}
        gate & flop & fail & flop & fail & flop & fail & flop & fail \\     
        \midrule
        0.75 & 0.65$\pm$0.07 & 0.00 & 0.80$\pm$0.06 & 0.00 & 0.15$\pm$0.04 & 0.00 & 2.00$\pm$0.00 & 0.00 \\ 
        0.5 & 0.66$\pm$0.04 & 0.00 & 0.88$\pm$0.06 & 0.00 & 0.02$\pm$0.01 & 0.00 & 1.00$\pm$0.00 & 0.00 \\ 
        0.25 & 1.05$\pm$0.04 & 0.00 & 1.83$\pm$0.06 & 0.25 & 0.02$\pm$0.01 & 0.00 & 1.00$\pm$0.00 & 0.00 \\ 
        0.1 & 1.82$\pm$0.05 & 0.00 & 3.08$\pm$0.16 & 0.00 & 0.04$\pm$0.02 & 0.00 & 1.00$\pm$0.00 & 0.00 \\ 
        0.05 & 2.44$\pm$0.08 & 0.00 & 3.81$\pm$0.06 & 0.50 & 0.05$\pm$0.03 & 0.00 & 1.00$\pm$0.00 & 0.00 \\
    \bottomrule
    \end{NiceTabular}
    \vskip 0.15in
    \begin{NiceTabular}{l cc cc cc cc}
    \toprule
        & \multicolumn{2}{c}{CCD} & \multicolumn{2}{c}{RCD} & \multicolumn{2}{c}{GCD} & \multicolumn{2}{c}{GD}\\ 
        \cmidrule(lr){2-3} \cmidrule(lr){4-5} \cmidrule(lr){6-7} \cmidrule(lr){8-9} 
        gate & flop & fail & flop & fail & flop & fail & flop & fail \\ 
        \midrule
        0.75 & 0.65$\pm$0.07 & 0.00 & 0.80$\pm$0.06 & 0.00 & 0.15$\pm$0.04 & 0.00 & 2.00$\pm$0.00 & 0.00 \\ 
        0.5 & 1.06$\pm$0.15 & 0.00 & 1.71$\pm$0.28 & 0.00 & 0.24$\pm$0.03 & 0.00 & 2.00$\pm$0.00 & 0.00\\ 
        0.25 & 1.75$\pm$0.13 & 0.00 & 3.28$\pm$0.29 & 0.25 & 0.54$\pm$0.27 & 0.00 & 2.44$\pm$0.51 & 0.00\\ 
        0.1 & 3.07$\pm$0.53 & 0.00 & - & 1.00 & 0.82$\pm$0.28 & 0.00 & 3.00$\pm$0.00 & 0.00 \\ 
        0.05 & 3.62$\pm$0.32 & 1.38 & - & 1.00 & 1.02$\pm$0.32 & 0.00 & 3.50$\pm$0.52 & 0.00 \\ 
    \bottomrule
    \end{NiceTabular}
    \end{scriptsize}
\end{table}

\subsection{Real datasets}
We further consider seven real-world datasets frequently used for assessing algorithm performance  in graph clustering (information can be found in the GitHub repository) \cite{mercado2019generalized,chodrow2021generative,venturini2022variance}:
\begin{itemize}
\item \textit{3sources}: 169 nodes, 6 communities, 3 layers;
\item \textit{BBCSport}: 544 nodes, 5 communities, 2 layers;
\item \textit{Wikipedia}: 693 nodes, 10 communities, 2 layers;
\item \textit{UCI}: 2000 nodes, 10 communities, 6 layers;
\item \textit{cora}: 2708 nodes, 7 communities, 2 layers;
\item \textit{primary-school}: 242 nodes, 11 communities, 2.4 mean hyperedge size;
\item \textit{high-school}: 327 nodes, 9 communities, 2.3 mean hyperedge size. 
\end{itemize}
The first five datasets in the list are related to multilayer graphs, while the last two are related to single layer hypergraphs.

We tested the methods considering different percentages of known labels per community, sampling them randomly 5 times and showing the average scores. In particular, we suppose to know $perc \in [3\%, 6\%, 9\%, 12\%]$ percentage of nodes per community in all the datasets except for \textit{Wikipedia}, where we considered to know a higher percentage of nodes, $perc \in [15\%, 18\%, 21\%, 24\%]$, to have significant results.

\begin{figure*}[ht]
\centering
{\includegraphics[scale=0.9,trim = 1.2cm 0.5cm 23cm 0.5cm, clip]{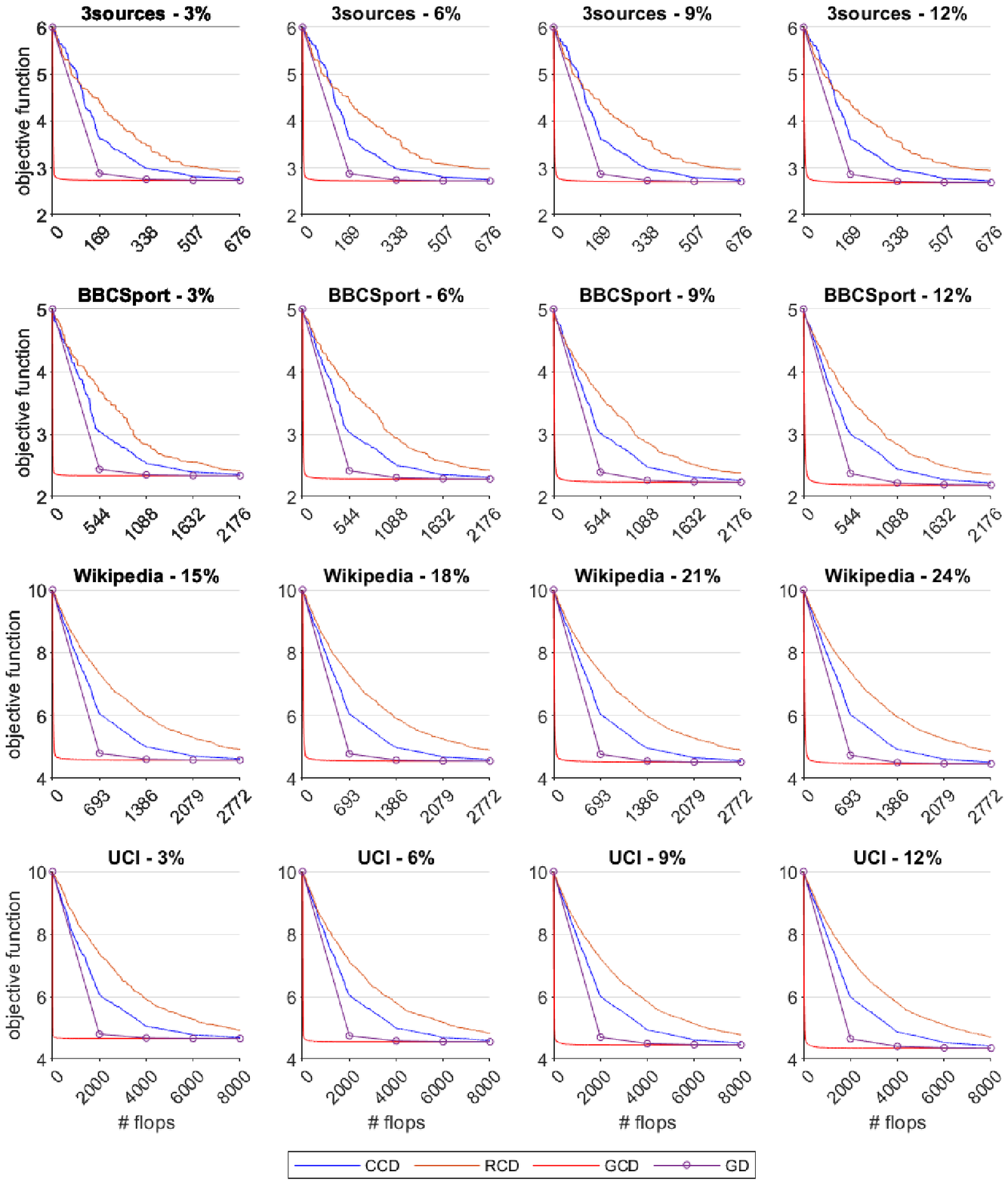}} 
\caption{
Average values of the objective function over 5 sampling of know labels, referring to 4 multilayer real-world datasets (3sources, BBCSport, Wikipedia, UCI) with quadratic regularizer. $perc \in [3\%, 6\%, 9\%, 12\%]$ (resp. $perc \in [15\%, 18\%, 21\%, 24\%]$ for Wikipedia) varies in the columns.}
\label{fig:real_fun_2_A}
\end{figure*}

\begin{figure*}[ht]
\centering
{\includegraphics[scale=0.9,trim = 1.2cm 0.5cm 23cm 0.5cm, clip]{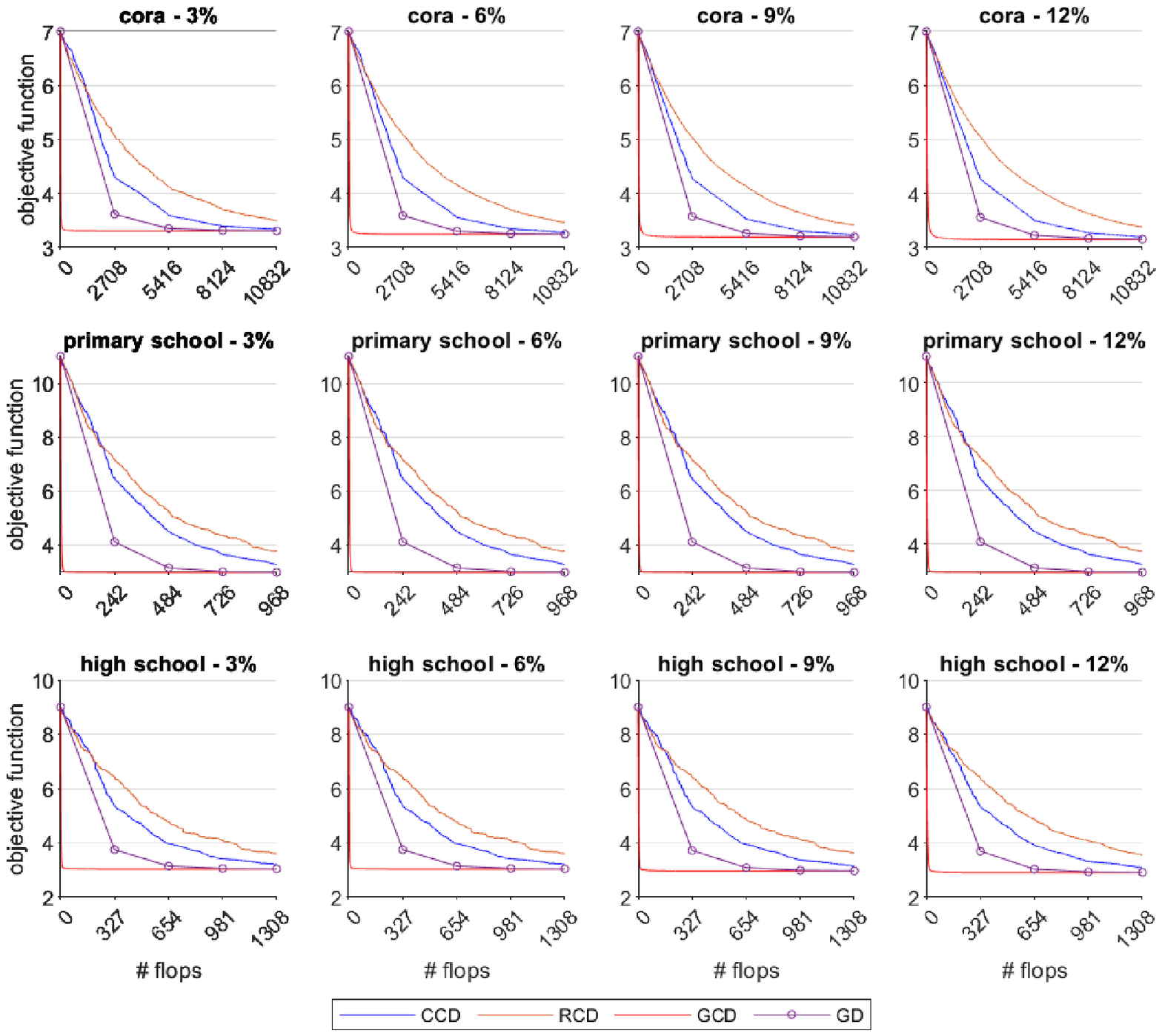}} 
\caption{
Average values of the objective function over 5 sampling of know labels, referring to 1 multilayer real-world dataset (cora) and 2 real-world hypergraphs  (primary school and high school) with quadratic regularizer. $perc \in [3\%, 6\%, 9\%, 12\%]$ varies in the columns.}
\label{fig:real_fun_2_B}
\end{figure*}

\begin{figure*}[ht]
\centering
{\includegraphics[scale=0.9,trim = 1cm 0.5cm 23cm 0.5cm, clip]{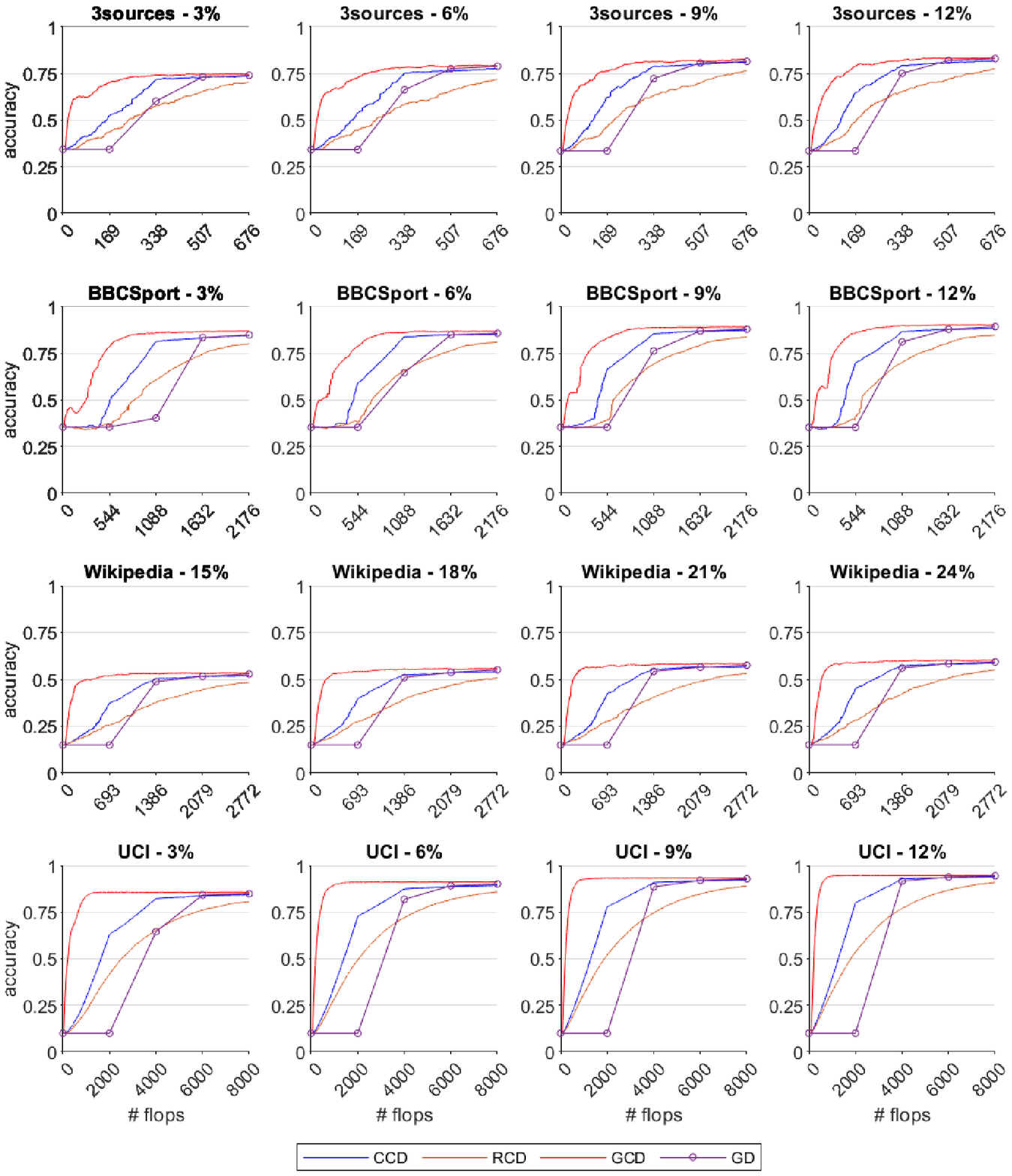}} 
\caption{
Average values of the accuracy over 5 sampling of know labels, referring to 4 multilayer real-world datasets (3sources, BBCSport, Wikipedia, UCI) with quadratic regularizer. $perc \in [3\%, 6\%, 9\%, 12\%]$ (resp. $perc \in [15\%, 18\%, 21\%, 24\%]$ for Wikipedia) varies in the columns.}
\label{fig:real_acc_2_A}
\end{figure*}

\begin{figure*}[ht]
\centering
{\includegraphics[scale=0.9,trim = 1cm 0.5cm 23cm 0.5cm, clip]{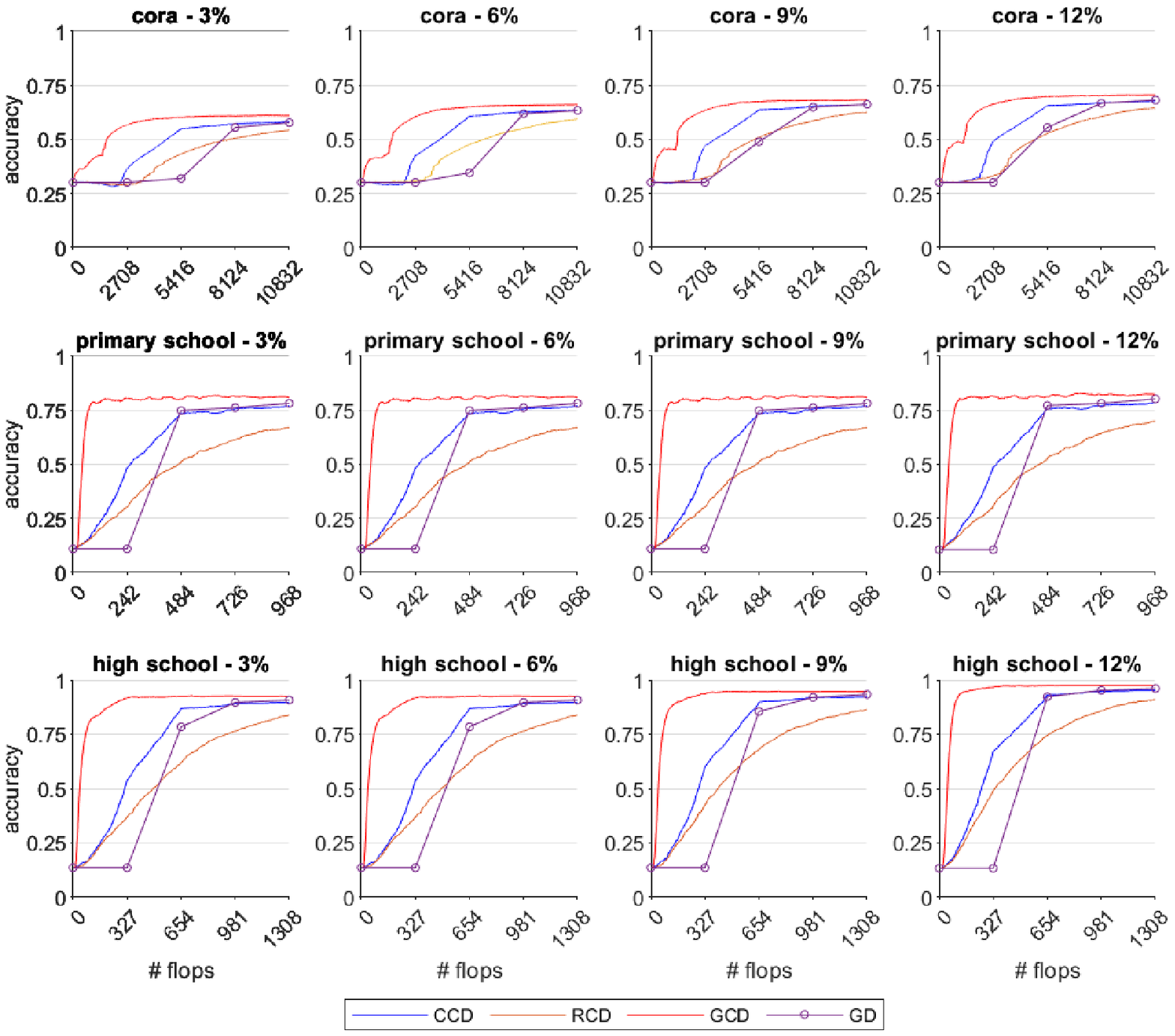}} 
\caption{
Average values of the accuracy over 5 sampling of know labels, referring to 1 multilayer real-world dataset (cora) and 2 real-world hypergraphs  (primary school and high school) with quadratic regularizer. $perc \in [3\%, 6\%, 9\%, 12\%]$ varies in the columns.}
\label{fig:real_acc_2_B}
\end{figure*}

Firstly, we analyze the results corresponding to a quadratic  regularization in~\eqref{prob_hypermultilayergraph} (fixing $p=2$). In  Figures \ref{fig:real_fun_2_A} and \ref{fig:real_fun_2_B}, we report the average values of the objective function, and in  Figures \ref{fig:real_acc_2_A} and \ref{fig:real_acc_2_B}, the related accuracy values. 
In Table \ref{tab:real_2}, we present aggregated results of the objective function and the accuracy, as explain in Section~\ref{Synthetic datasets}.
We can see that the results match the ones obtained for the synthetic datasets. The Greedy Coordinate Descent method (GCD) indeed reaches a good solution in terms of both objective function and accuracy, with a much lower number of flops than the other methods.

\setlength{\tabcolsep}{2.5pt}
\begin{table}[t]
\caption{
Aggregated results of the objective function (upper table) and the accuracy (lower table) across the real datasets with $p=2$ (see Figures \ref{fig:real_fun_2_A}-\ref{fig:real_fun_2_B} and Figures \ref{fig:real_acc_2_A}-\ref{fig:real_acc_2_B}).
The table indices are the same as in Table \ref{tab:art}.
}
\label{tab:real_2}
    \centering
    \begin{scriptsize}
    \vskip 0.15in
    \begin{NiceTabular}{l cc cc cc cc}
    \toprule
        & \multicolumn{2}{c}{CCD} & \multicolumn{2}{c}{RCD} & \multicolumn{2}{c}{GCD} & \multicolumn{2}{c}{GD}\\ 
        \cmidrule(lr){2-3} \cmidrule(lr){4-5} \cmidrule(lr){6-7} \cmidrule(lr){8-9}
        gate & flop & fail & flop & fail & flop & fail & flop & fail \\     
        \midrule
        0.75 & 0.39$\pm$ 0.06& 0.00 & 0.40$\pm$0.02 & 0.00 & 0.01$\pm$0.00 & 0.00 & 1.00$\pm$0.00 & 0.00 \\ 
        0.5 & 0.74$\pm$0.08 & 0.00 & 1.06$\pm$0.09 & 0.00 & 0.01$\pm$0.00 & 0.00 & 1.00$\pm$0.00 & 0.00 \\ 
        0.25 & 1.30$\pm$0.24 & 0.00 & 2.06$\pm$0.16 & 0.00 & 0.02$\pm$0.01 & 0.00 & 1.00$\pm$0.00 & 0.00 \\ 
        0.1 & 2.16$\pm$0.40 & 0.00 & 3.37$\pm$0.36 & 0.07 & 0.03$\pm$0.01 & 0.00 & 1.32$\pm$0.48 & 0.00 \\ 
        0.05 & 2.90$\pm$0.49 & 0.00 & 3.49$\pm$0.00 & 0.96 & 0.04$\pm$0.02 & 0.00 & 1.61$\pm$0.50 & 0.00 \\ 
    \bottomrule
    \end{NiceTabular}
    \vskip 0.15in
    \begin{NiceTabular}{l cc cc cc cc}
    \toprule
        & \multicolumn{2}{c}{CCD} & \multicolumn{2}{c}{RCD} & \multicolumn{2}{c}{GCD} & \multicolumn{2}{c}{GD}\\ 
        \cmidrule(lr){2-3} \cmidrule(lr){4-5} \cmidrule(lr){6-7} \cmidrule(lr){8-9}
        gate & flop & fail & flop & fail & flop & fail & flop & fail \\  
        \midrule
        0.75 & 0.69$\pm$0.17 & 0.00 & 0.97$\pm$0.28 & 0.00 & 0.13$\pm$0.08 & 0.00 & 2.11$\pm$0.31 & 0.00 \\ 
        0.5 & 0.99$\pm$0.20 & 0.00 & 1.61$\pm$0.31 & 0.00 & 0.25$\pm$0.16 & 0.00 & 2.15$\pm$0.36 & 0.00 \\ 
        0.25 & 1.51$\pm$0.26 & 0.00 & 2.78$\pm$0.49 & 0.00 & 0.41$\pm$0.24 & 0.00 & 2.32$\pm$0.48 & 0.00 \\ 
        0.1 & 2.15$\pm$0.52 & 0.00 & 3.21$\pm$0.23 & 0.82 & 0.67$\pm$0.39 & 0.00 & 2.71$\pm$0.54 & 0.04 \\ 
        0.05 & 2.70$\pm$0.56 & 0.29 & 3.86$\pm$0.00 & 0.96 & 0.91$\pm$0.52 & 0.00 & 3.10$\pm$0.44 & 0.25 \\
    \bottomrule
    \end{NiceTabular}
    \end{scriptsize}
\end{table}

\begin{figure*}[ht]
\centering
{\includegraphics[scale=0.9,trim = 1cm 0.5cm 23cm 0.5cm, clip]{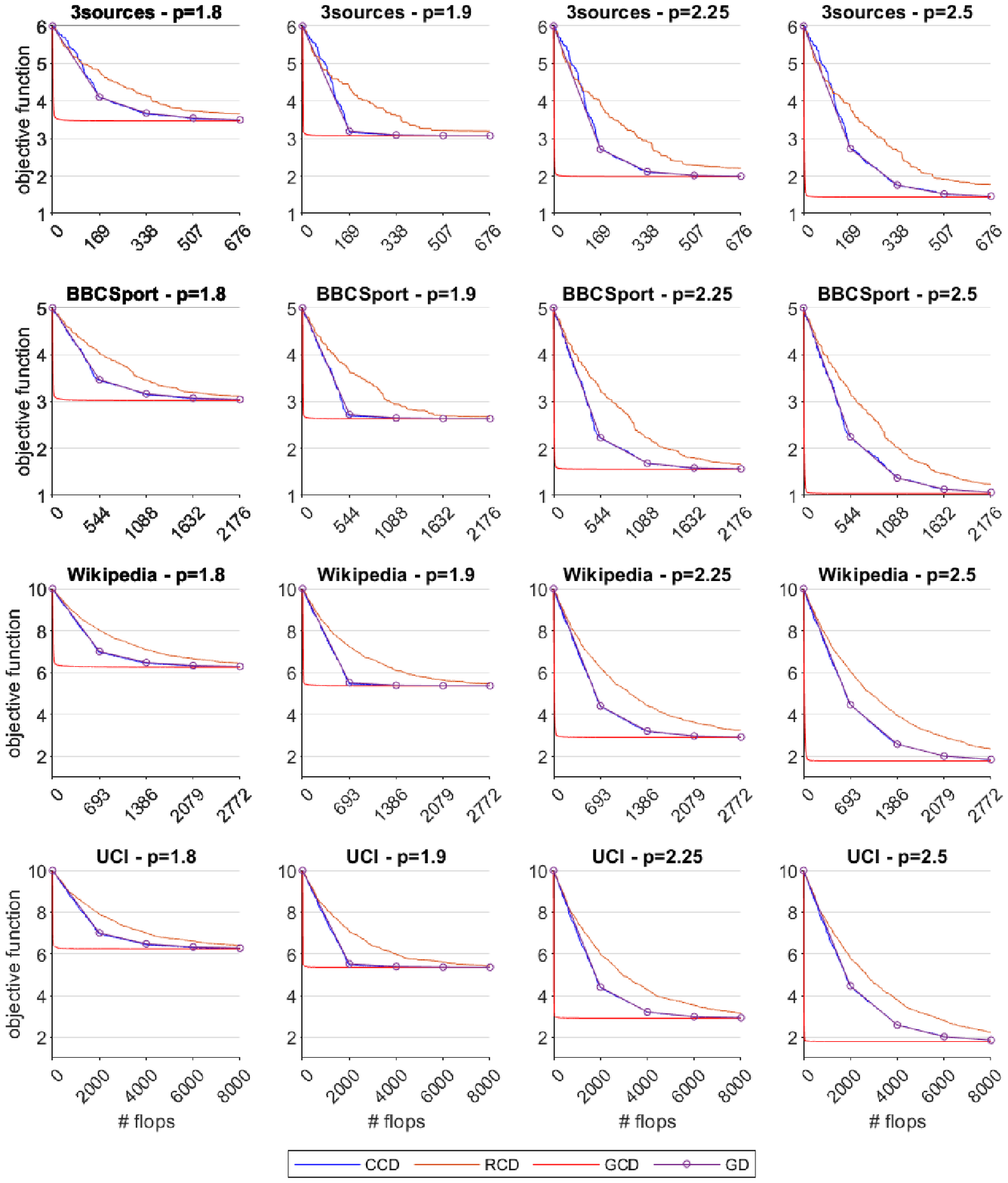}} 
\caption{
Average values of the objective function over 5 sampling of know labels, referring to 4 multilayer real-world datasets (3sources, BBCSport, Wikipedia, UCI) with $perc=6\%$ (resp. $perc=18\%$ for Wikipedia). $p \in [1.8,1.9,2.25,2.5]$ in the regularization term varies in the columns.}
\label{fig:real_fun_p_A}
\end{figure*}

\begin{figure*}[ht]
\centering
{\includegraphics[scale=0.9,trim = 1cm 0.5cm 23cm 0.5cm, clip]{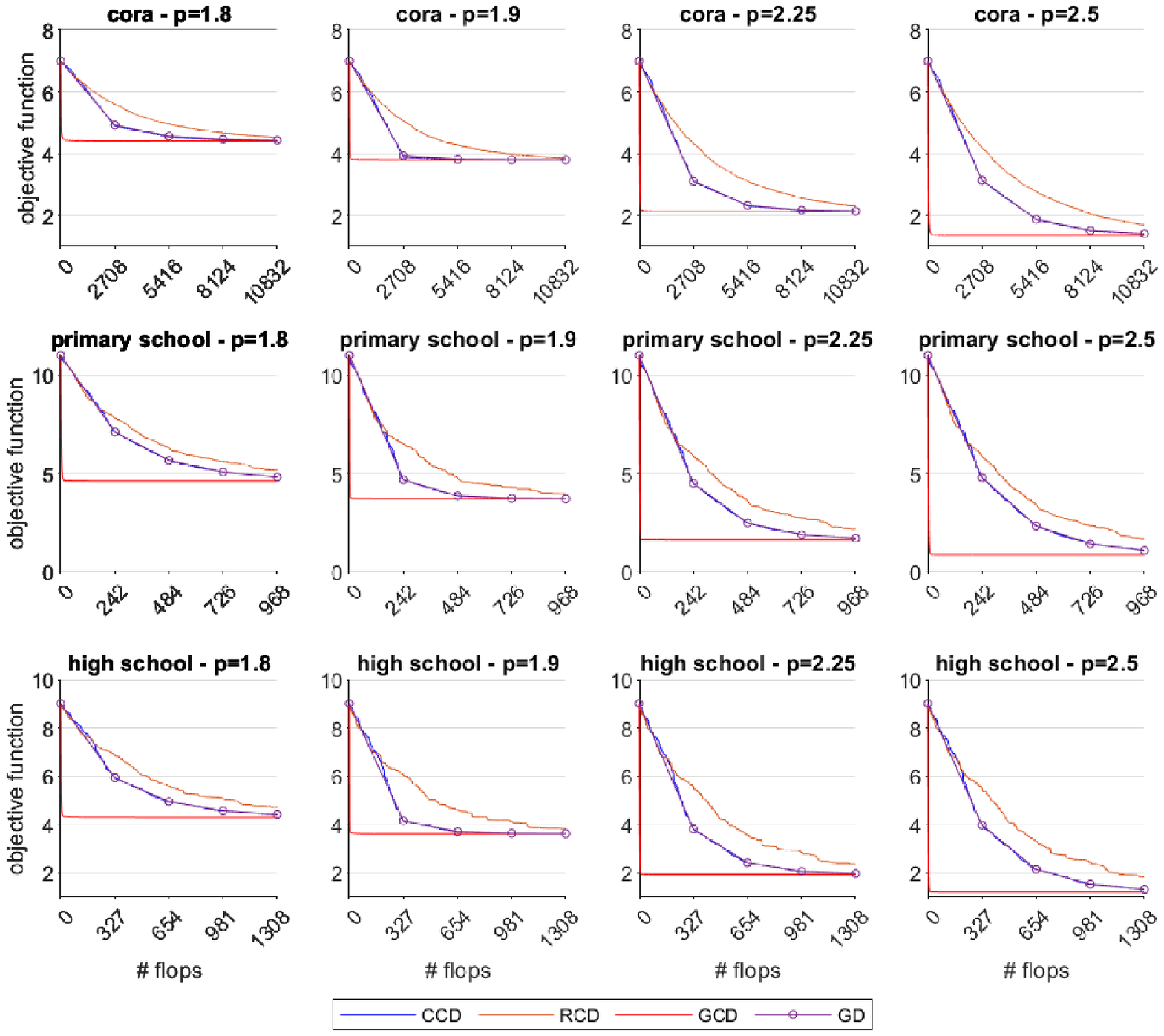}} 
\caption{
Average values of the objective function over 5 sampling of know labels, referring to 1 multilayer real-world dataset (cora) and 2 real-world hypergraphs  (primary school and high school) with $perc=6\%$. $p \in [1.8,1.9,2.25,2.5]$ in the regularization term varies in the columns.}
\label{fig:real_fun_p_B}
\end{figure*}

\begin{figure*}[ht]
\centering
{\includegraphics[scale=0.9,trim = 1cm 0.5cm 23cm 0.5cm, clip]{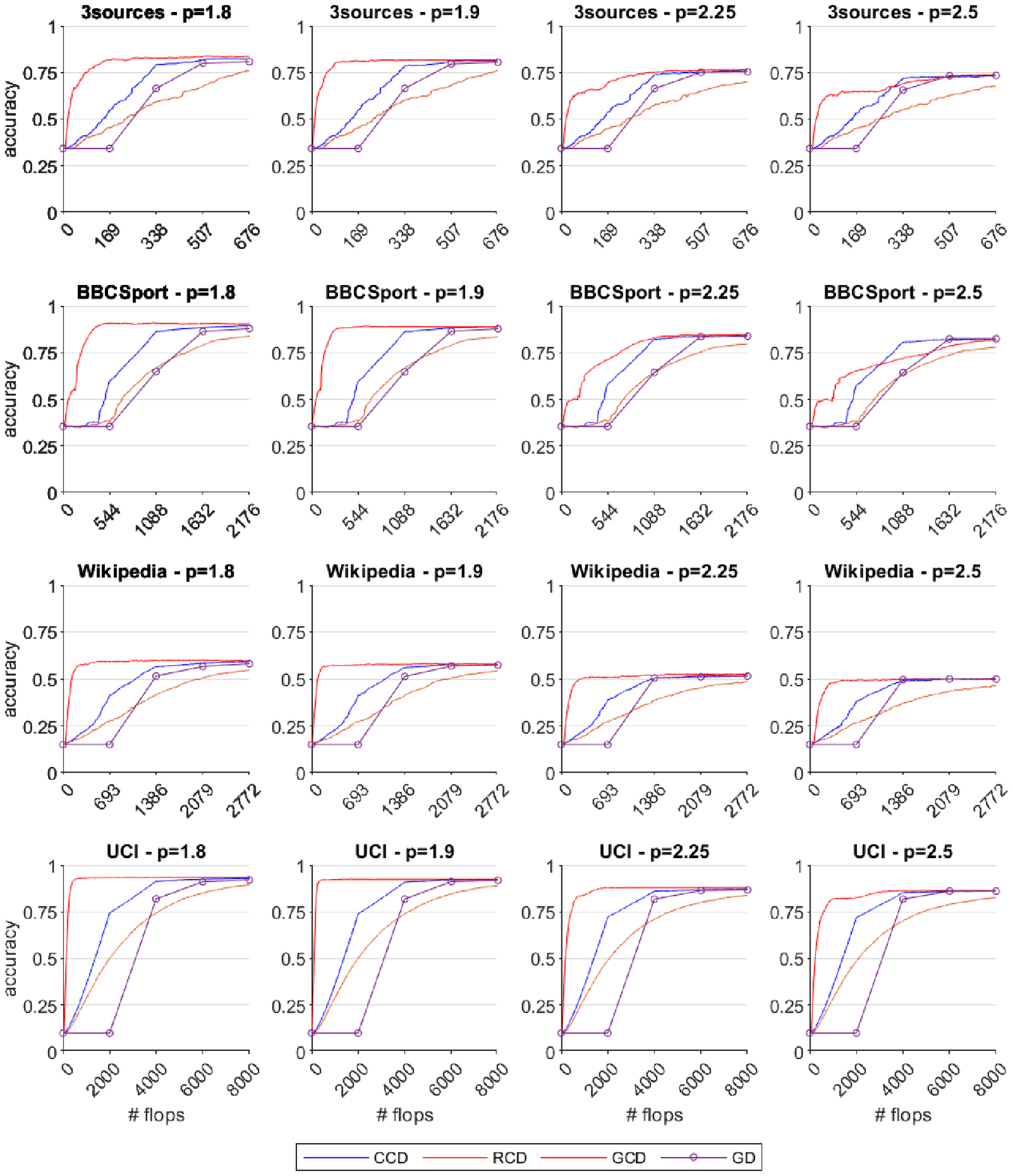}} 
\caption{
Average values of the accuracy over 5 sampling of know labels, referring to 4 multilayer real-world datasets (3sources, BBCSport, Wikipedia, UCI) with $perc=6\%$ (resp. $perc=18\%$ for Wikipedia). $p \in [1.8,1.9,2.25,2.5]$ in the regularization term varies in the columns.}
\label{fig:real_acc_p_A}
\end{figure*}

\begin{figure*}[ht]
\centering
{\includegraphics[scale=0.9,trim = 1cm 0.5cm 23cm 0.5cm, clip]{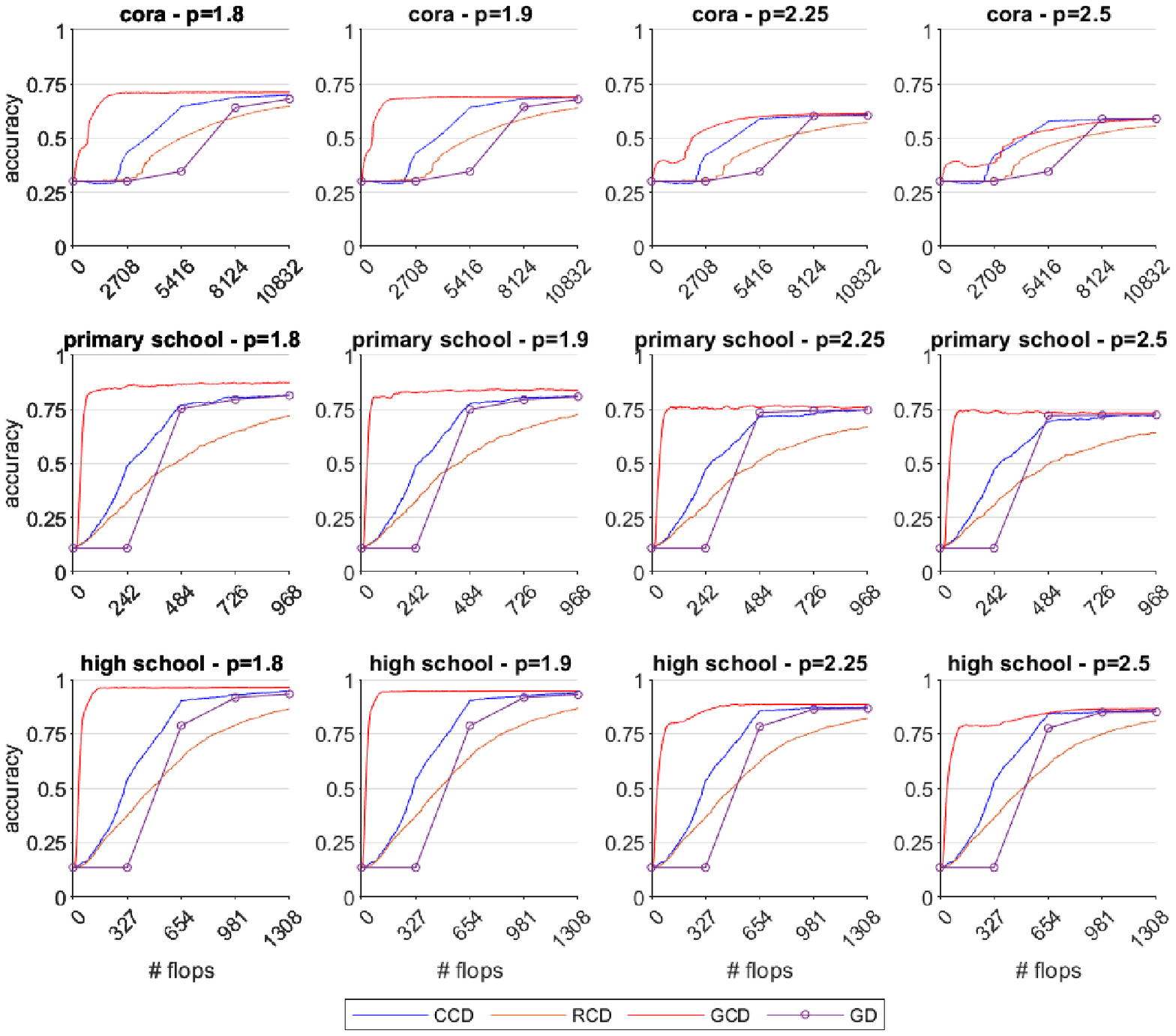}} 
\caption{
Average values of the accuracy over 5 sampling of know labels, referring to 1 multilayer real-world dataset (cora) and 2 real-world hypergraphs  (primary school and high school) with $perc=6\%$. $p \in [1.8,1.9,2.25,2.5]$ in the regularization term varies in the columns.}
\label{fig:real_acc_p_B}
\end{figure*}

In order to investigate how the regularization parameter $p$  influences the behavior of the methods and the accuracy of the results,  we further carried out experiments with $p\neq2$. In particular, we took into consideration both values larger and smaller than $2$. We show the results for $p \in \{1.8,1.9,2.25,2.5\}$, with fixed $perc=6\%$ (resp. \textit{Wikipedia} $perc=18\%$).
In  Figures \ref{fig:real_fun_p_A} and \ref{fig:real_fun_p_B}, we report the average values of the objective function, and in  Figures \ref{fig:real_acc_p_A} and \ref{fig:real_acc_p_B}, the related accuracy values. 
In Table \ref{tab:real_p}, we present aggregated results of the objective function and the accuracy, as explain in Section~\ref{Synthetic datasets}.
We can notice that the behavior of the methods does not change much by varying $p$. The GCD method performs better than the others in terms of number of flops. Looking at the values of the objective function, the CCD method seems to have a behavior very similar to the ones of the GD. Meanwhile, the RCD method performs poorly both in terms of objective function and accuracy. It is important to observe that $p\neq2$ can lead to an improvement of the final accuracy. In Table \ref{tab:real_p_acc}, we report the maximum value of accuracy achieved in the real datasets with fixed $perc =6\%$ (resp. $perc = 18\%$ for Wikipedia) and varing $p \in \{1.8, 1.9, 2, 2.25, 2.5\}$.

\setlength{\tabcolsep}{2.5pt}
\begin{table}[t]
\caption{
Aggregated results of the objective function (upper table) and the accuracy (lower table) across the real datasets with $p\neq2$ (see Figures \ref{fig:real_fun_p_A}-\ref{fig:real_fun_p_B} and Figures \ref{fig:real_acc_p_A}-\ref{fig:real_acc_p_B}).
The table indices are the same as in Table \ref{tab:art}.
}
\label{tab:real_p}
    \centering
    \begin{scriptsize}
    \vskip 0.15in
    \begin{NiceTabular}{l cc cc cc cc}
    \toprule
        & \multicolumn{2}{c}{CCD} & \multicolumn{2}{c}{RCD} & \multicolumn{2}{c}{GCD} & \multicolumn{2}{c}{GD}\\ 
        \cmidrule(lr){2-3} \cmidrule(lr){4-5} \cmidrule(lr){6-7} \cmidrule(lr){8-9}
        gate & flop & fail & flop & fail & flop & fail & flop & fail \\      
        \midrule
        0.75 & 0.35$\pm$0.06 & 0.00 & 0.35$\pm$0.05 & 0.00 & 0.01$\pm$0.00 & 0.00 & 1.00$\pm$0.00 & 0.00 \\ 
        0.5 & 0.66$\pm$0.09 & 0.00 & 0.93$\pm$0.14 & 0.00 & 0.01$\pm$0.00 & 0.00 & 1.00$\pm$0.00 & 0.00 \\ 
        0.25 & 1.06$\pm$0.26 & 0.00 & 1.83$\pm$0.21 & 0.00 & 0.01$\pm$0.01 & 0.00 & 1.43$\pm$0.50 & 0.00 \\ 
        0.1 & 1.68$\pm$0.49 & 0.00 & 2.99$\pm$0.43 & 0.00 & 0.02$\pm$0.01 & 0.00 & 1.93$\pm$0.60 & 0.00 \\ 
        0.05 & 2.13$\pm$0.72 & 0.00 & 3.40$\pm$0.39 & 0.50 & 0.03$\pm$0.01 & 0.00 & 2.50$\pm$0.29 & 0.00 \\ 
    \bottomrule
    \end{NiceTabular}
    \vskip 0.15in
    \begin{NiceTabular}{l cc cc cc cc}
    \toprule
        & \multicolumn{2}{c}{CCD} & \multicolumn{2}{c}{RCD} & \multicolumn{2}{c}{GCD} & \multicolumn{2}{c}{GD}\\ 
        \cmidrule(lr){2-3} \cmidrule(lr){4-5} \cmidrule(lr){6-7} \cmidrule(lr){8-9}
        gate & flop & fail & flop & fail & flop & fail & flop & fail \\  
        \midrule
        0.75 & 0.70$\pm$0.16 & 0.00 & 0.97$\pm$0.28 & 0.00 & 0.09$\pm$0.03 & 0.00 & 2.15$\pm$0.36 & 0.00 \\ 
        0.5 & 1.01$\pm$0.19 & 0.00 & 1.64$\pm$0.29 & 0.00 & 0.23$\pm$0.24 & 0.00 & 2.15$\pm$0.36 & 0.00 \\ 
        0.25 & 1.51$\pm$0.26 & 0.00 & 2.72$\pm$0.45 & 0.00 & 0.41$\pm$0.45 & 0.00 & 2.36$\pm$0.49 & 0.00 \\ 
        0.1 & 1.93$\pm$0.26 & 0.00 & 3.46$\pm$0.46 & 0.68 & 0.72$\pm$0.76 & 0.00 & 2.90$\pm$0.57 & 0.00 \\ 
        0.05 & 2.39$\pm$0.54 & 0.04 & 3.90$\pm$0.11 & 0.89 & 0.94$\pm$0.90 & 0.00 & 3.05$\pm$0.56 & 0.18 \\ 
    \bottomrule
    \end{NiceTabular}
    \end{scriptsize}
\end{table}

\setlength{\tabcolsep}{3pt}
\begin{table}[t]
\caption{
Maximum value of accuracy achieved in the real datasets with fixed $perc =6\%$ (resp. $perc = 18\%$ for Wikipedia) and varing $p \in \{1.8, 1.9, 2, 2.25, 2.5\}$. 
}
\label{tab:real_p_acc}
    \centering
    \begin{scriptsize}
    \vskip 0.15in
    \begin{NiceTabular}{l ccccc}
    \toprule
    & \multicolumn{5}{c}{$p$}\\ 
    \cmidrule(lr){2-6}
    dataset & 1.8 & 1.9 & 2 & 2.25 & 2.5 \\
    \midrule
    3sources & 0.84 & 0.82 & 0.79 & 0.76 & 0.74 \\ 
    BBCSport & 0.91 & 0.89 & 0.87 & 0.85 & 0.83 \\ 
    Wikipedia & 0.60 & 0.58 & 0.56 & 0.53 & 0.50 \\ 
    UCI & 0.94 & 0.93 & 0.91 & 0.88 & 0.86 \\ 
    cora & 0.71 & 0.69 & 0.66 & 0.62 & 0.60 \\ 
    primary school & 0.89 & 0.85 & 0.82 & 0.77 & 0.75 \\ 
    high school & 0.96 & 0.95 & 0.93 & 0.89 & 0.87 \\ 
    \bottomrule
    \end{NiceTabular}
    \end{scriptsize}
\end{table}

\section{Conclusions}
\label{Conclusions}
In this paper, we compared different coordinate descent methods with the standard Gradient Descent approach for the resolution of an optimization-based formulation of the Graph Semi-Supervised learning problem on multilayer hypergraphs. 
We performed extensive experiments on both synthetic and real world datasets, which show the faster convergence speed of suitably chosen coordinate methods with respect to the Gradient Descent approach. This fact clearly indicates that the design of tailored coordinate methods for the resolution of the considered semi-supervised learning problems  represents a fruitful path to follow. 
In addition, we carried out an analysis replacing the standard quadratic regularization term in the objective function with a more general $p-$regularizer. The reported results clearly show that this modification can lead to better performances. 

\appendix
\section{Calculations}
\label{appendix:A}
In order to compare the gradient descent method to block coordinate descent approaches, we need to calculate the gradient of the function $\vartheta(Z)$ that we want to minimize in~\eqref{prob_hypermultilayergraph}. 
%and the stepsize $\alpha^k$.
The gradient of $\vartheta(Z)$ can be expressed as:
\begin{equation}
\label{grad}
\nabla \vartheta(Z) = 2(Z - Y) +  p\sum_{\ell=1}^L \lambda_{\ell} {\mathcal{L}_{\ell}^p}(Z),
\end{equation}
where ${\mathcal{L}_{\ell}^p}(Z)$ is the normalized $p-$laplacian and it is applied on each column of $Z$ in this way:
\[
\mathcal{L}_{\ell}^p(Z) = ( B_{\ell}D_\ell^{-\frac{1}{2}})^T \, \phi_p (B_{\ell} D_{\ell}^{-\frac{1}{2}} Z),
\]
 with $\phi_p (y) = |y|^{p-1} \mathrm{sgn}(y)$ component-wise.

At the beginning, we calculate $B_{\ell} D_{\ell}^{-\frac{1}{2}}$ $\forall \ell \in \{1,..,L\}$ layer.
Then, at each iteration $k$, the gradient is calculated in an iterative fashion. Break the formula of the gradient in~\eqref{grad} into two parts:
\begin{equation*}
    \nabla \vartheta(Z) = \nabla f(Z) + \nabla r_p(Z),
\end{equation*}
with 
\begin{align*}
    \nabla f(Z) & = 2(Z - Y), \\
    \nabla r_p(Z) & = p\sum_{\ell=1}^L \lambda^{\ell} {\mathcal{L}_{\ell}^p}(Z).
\end{align*}
Then, 
\begin{align*}
    \nabla f(Z^{k+1}) & = \nabla f(Z^{k}) + 2(Z^{K+1}_{W^k} - Z^{K}_{W^k}), \\
    \nabla r_p(Z^{k+1}) & =  p\sum_{\ell=1}^L \lambda^{\ell} {\mathcal{L}_{\ell}^p}(Z^{k+1}) ,
\end{align*}
where ${\mathcal{L}_{\ell}^p}(Z^{k+1})$ can be iteratively calculated using 
\[
B_{\ell} D_{\ell}^{-\frac{1}{2}} Z^{K+1} = B_{\ell} D_{\ell}^{-\frac{1}{2}} Z^{K} + (B_{\ell} D_{\ell}^{-\frac{1}{2}})_{W^k} (Z^{K+1}_{W^k} - Z^{K}_{W^k})
\]
with the appropriate subscript $W^k$ to take just the $W_j^k$ coordinates of column $j$, for all $j \in \{ 1,\ldots, m\}$.
% We use as step size, both for the gradient descent algorithm and the coordinate descent methods, 
% \[
% \alpha^k = \frac{1}{2(2 + \lambda p 2^{\frac{p-1}{p}})} \, .
% \]
% This choice ensures convergence of the algorithms, since by \cite{amghibech2006bounds} it directly follows that 
% \begin{theorem}
% If $G$ is a connected graph, then
% \begin{equation}
% \lambda \leq 2^{p - 1} %max\{d(v), v \in V \}
% \end{equation}
% for any $\lambda$ eigenvalue of the $p$-Laplacian operator $\mathcal L_p^\ell$. 
% \end{theorem}
% Then, from 
% \begin{equation}
%     \lVert  \tilde{\mathcal{L}_p} (x) -  \tilde{\mathcal{L}_p} (y) \rVert_{p}^p \leq
%     2^{p-1} \lVert x - y \rVert_{p}^p 
% \end{equation}
% follows that 
% \begin{equation}
%     \lVert \nabla \tilde{\mathcal{L}_p} (x) - \nabla \tilde{\mathcal{L}_p} (y) \rVert_{p*}^p \leq 
%     2^{p-1} \lVert x - y \rVert_{p}^p 
% \end{equation}
% where $p>1$ and $p*$ dual of $p$.
% The convergence is ensured by theorem  

\subsection{Special case $p=2$}
In this section, we discuss the special case of problem~\eqref{prob_hypermultilayergraph} with $p=2$.
The optimization problem~\eqref{prob_hypermultilayergraph} is equivalent to:
\[
\min_{Z \in \R^{|V| \times k}} \lVert  Z -Y \rVert_{(2)}^2 +  \sum_{\ell=1}^L \lambda_{\ell} Z^{T} \bar L_{\ell} Z,
\]
where $\bar L_{\ell} = I-\bar A_{\ell}$ is the normalized laplacian matrix of layer $\ell=1,\ldots,L$ and
$\bar A_{\ell}$ is the normalized adjacency matrix of layer $\ell=1,\ldots,L$ with entries 
\[
 (\bar A_{\ell})_{uv} = \frac{(A_{\ell})_{uv}}{\sqrt{(\delta_{\ell})_u}\sqrt{(\delta_{\ell})_v}}.
\]
In this case, the gradient of the function to minimize can be expressed as
\[
\nabla_Z \vartheta(Z) = 2(Z - Y) + \sum_{\ell=1}^L 2\lambda_{\ell} \bar L_{\ell} Z 
\]
and the Hessian as $2I + \sum_{\ell=1}^L 2\lambda_{\ell} \bar L_{\ell}$.
In the experiments where $p=2$, this last expression can be used in the calculation of the stepsize.

\clearpage

\printbibliography

\end{document}